\pdfoutput=1

\documentclass[11pt]{article}

\usepackage{EMNLP2023}

\usepackage{times}
\usepackage{latexsym}

\usepackage[T1]{fontenc}

\usepackage[utf8]{inputenc}

\usepackage{microtype}

\usepackage{inconsolata}

\usepackage[final]{changes}  
\definechangesauthor[name=Viktor, color=blue]{VH}
\definechangesauthor[name=Silvia, color=orange]{S}

\usepackage{algorithm}
\usepackage{algpseudocode}
\usepackage{graphicx}
\usepackage{booktabs}
\usepackage{xspace}
\usepackage{amsmath}
\usepackage{multirow}
\usepackage{numprint}

\title{Multilingual Word Embeddings for Low-Resource Languages \\ using Anchors and a Chain of Related Languages}

\author{Viktor Hangya$^{1,2}$, Silvia Severini$^1$, Radoslav Ralev$^3$, \\
\textbf{Alexander Fraser}$^{1,2}$ \and \textbf{Hinrich Sch{\"u}tze}$^{1,2}$ \\
  $^1$Center for Information and Language Processing, LMU Munich, Germany \\
  $^2$Munich Center for Machine Learning, \\
  $^3$Technical University of Munich \\
  {\tt \{hangyav, silvia, fraser\}@cis.lmu.de}, \\ {\tt radoslav.ralev@tum.de}
}

\begin{document}
\maketitle

\newcommand{\ourmethod}{\textsc{ChainMWEs}\xspace} \newcommand{\anchors}{\textsc{AnchorBWEs}\xspace} 

\begin{abstract}

Very low-resource languages,
having only a few million tokens worth of data,
are not well-supported by multilingual NLP
approaches due to
poor quality
cross-lingual word representations.
Recent work showed that good cross-lingual performance can be
achieved if a source language is related to the low-resource target language.
However, not all language pairs are related.
In this paper, we propose to
build multilingual word embeddings (MWEs)
via a novel language chain-based approach, that incorporates intermediate
related languages to bridge the gap between the distant source
and
target.
We build MWEs one language at a time by starting from the
resource rich source
and sequentially adding each language in the chain
till we reach the target.
We
extend a semi-joint bilingual approach to
multiple languages in order to eliminate the main weakness of previous works,
i.e., independently trained monolingual embeddings, by
anchoring the target language around the
multilingual space.
We evaluate our method on bilingual lexicon induction for 4
language families, involving 4 very low-resource ($\le$ 5M tokens) and 4
moderately low-resource ($\le$ 50M)
target languages,
showing improved performance in both categories.
Additionally, our analysis reveals the importance of good quality embeddings for
intermediate languages
as well as
the importance of leveraging anchor points from all languages in the
multilingual space.

\end{abstract}

\section{Introduction}
\label{sec:intro}

Cross-lingual word representations are shared embedding spaces for two --
\textit{Bilingual (BWEs)} -- or more languages -- \textit{Multilingual Word Embeddings (MWEs)}.
They have been shown to be effective for multiple tasks including machine translation \cite{lample-etal-2018-phrase} and cross-lingual transfer learning \cite{schuster-etal-2019-cross}.
They can be created by jointly learning shared embedding spaces
\cite{Lample2018,conneau-etal-2020-unsupervised} or via mapping approaches
\cite{artetxe-etal-2018-robust,schuster-etal-2019-cross}.
However, their quality degrades when low-resource languages are involved,
since they require an adequate amount of monolingual data
\citep{adams-etal-2017-cross}, which is especially problematic
for languages with just a few millions of tokens
\cite{eder2021anchorbased}.

Recent work showed that building embeddings jointly by
representing common vocabulary items of the source and target languages with a
single embedding can improve representations~\cite{wang2019cross,woller-etal-2021-neglect}.
On the other hand, these approaches require the source and target to
be related, which in practice means high vocabulary overlap.
Since for many distant language pairs this requirement is not satisfied,
in this paper, we
propose to leverage a chain of intermediate languages to overcome the large
language gap.
We build MWEs step-by-step, starting from the source language and moving towards the
target, incorporating a language that is related to the languages
already in the multilingual space in each step.
Intermediate languages are selected based on their linguistic proximity
to the source and target languages, as well as the availability of large enough
datasets.

Since our main targets are
languages having
just a few million tokens worth of monolingual data, we take static word embeddings
\cite{mikolov2013efficient} instead of contextualized representations
\cite{devlin-etal-2019-bert} as the basis of our method, due to the generally
larger data requirements
of the latter.
Additionally, the widely used mapping-based approaches \cite{mikolov2013exploiting}, including
multilingual methods
\cite{kementchedjhieva-etal-2018-generalizing,jawanpuria2018learning,chen-cardie-2018-unsupervised},
require good quality monolingual word embeddings.
Thus, to incorporate a single language to the multilingual space in each step
we rely on the anchor-based approach of \newcite{eder2021anchorbased}.
We refer to this method as \anchors.
It builds the target embeddings and aligns them to the source space in one step
using anchor points, thus not only building cross-lingual representations but a
better quality target language space as well.
We extend this bilingual approach to multiple languages.
Instead of aligning the target language to the source in one step, we maintain a
multilingual space (initialized by the source language), and adding each
intermediate and finally the target language to it sequentially.
This way we make sure that the language gap between the two spaces in each step
stays minimal.

We evaluate our approach (\ourmethod) on the Bilingual Lexicon Induction (BLI)
task for 4
language families, including 4 very ($\le$ 5 million tokens) and 4 moderately
low-resource ($\le$ 50 million) languages and
show improved performance compared to both bilingual and multilingual
mapping based baselines, as well as to the bilingual \anchors.
Additionally, we analyze the importance of intermediate language quality, as well
as the role of the number of anchor points during training.
In summary, our contributions are the following:

\begin{itemize}
    \item we propose to strengthen word embeddings of low-resource
    languages by employing a chain of intermediate related languages in order to
    reduce the language gap at each alignment step,
    \item we extend \anchors of \newcite{eder2021anchorbased}
    to multilingual word representations which does not take the distance
    between the source and target languages into consideration,
        \item we test our approach on multiple low-resource languages and show
    improved performance,
    \item we make our code available for public use.\footnote{\url{https://cistern.cis.lmu.de/anchor-embeddings}}
\end{itemize}

\section{Related Work}
\label{sec:previouswork}

Bilingual lexicon induction is the task of inducing word translations from
monolingual corpora in two languages
\citep{irvine-callison-burch-2017-comprehensive},
which became the de facto task to evaluate the quality of cross-lingual word
embeddings.
There are two main approaches to obtain MWEs: mapping and joint learning.
Mapping approaches aim at computing a transformation matrix to map the embedding
space of one language onto the embedding space of the others
\citep[inter alia]{ravi-knight-2011-deciphering, artetxe-etal-2017-learning, conneau2018word, artetxe-etal-2018-robust,Lample2018,artetxe-etal-2019-bilingual}.
Alternatively, joint learning approaches aim at learning a shared embedding
space for two or more languages simultaneously.
 \citet{luong2015bilingual} learn sentence and word-level alignments jointly and
create BWEs by modifying the Skip-gram model. The Skip-gram model is also used
by \citet{vulic2015bilingual}  who train it on a pseudo-bilingual corpus obtained by merging two aligned documents.
\citet{artetxe2019massively} use a large parallel corpus to train a bidirectional LSTM and jointly learn representations for many languages.
Most recently, transformer based large LMs are trained jointly on multiple
languages using a shared subword vocabulary to obtain contextualized
cross-lingual representations
\cite{devlin-etal-2019-bert,conneau-etal-2020-unsupervised}.
However, large LMs require more training data than static word embeddings,
thus we focus on the latter in our work.

\citet{Ruder_2019} provided a survey paper on cross-lingual word
embedding models and identified three sub-categories within static word-level
alignment models: mapping-based approaches, pseudo-multilingual corpus-based
approaches and joint methods, highlighting their advantages and disadvantages.
To combine the advantages of mapping and joint approaches \citet{wang2019cross}
proposed to first apply joint training followed by a mapping step on overshared
words, such as false friends.
Similarly, a hybrid approach was introduced in \cite{woller-etal-2021-neglect}
for 3 languages, which first applies joint training on two related languages
which is then mapped to the distant third language.
A semi-joint approach was introduced in \cite{ormazabal-etal-2021-beyond} and
\cite{eder2021anchorbased}, which using a fixed pre-trained monolingual space
of the source language trains the target space from scratch by aligning
embeddings close to given source anchor points.
We utilize \citep{eder2021anchorbased}
in our work, since it is evaluated on very low-resource languages which is the
main interest of our work.

Most work on cross-lingual word embeddings is English-centric.
\citet{Anastasopoulos2019} found that the choice of hub language to which others
are aligned to can significantly affect the final performance.
Other methods leveraged multiple languages to build MWEs
\cite{kementchedjhieva-etal-2018-generalizing,chen-cardie-2018-unsupervised,jawanpuria2018learning},
showing that some languages can help each other to achieve improved performance
compared to bilingual systems.
However, these approaches rely on pre-trained monolingual embeddings, which
could be difficult to train in limited resource scenarios.
In our work we also leverage multiple languages, but mitigate the issue of poor
quality monolingual embeddings.

\citet{sogaard-etal-2018-limitations}
showed that embedding spaces do not tend to be
isomorphic in case of distant or low-resource language pairs, making the task of aligning
monolingual word
embeddings harder than previously assumed.
Similarly, \citet{patra-etal-2019-bilingual} empirically show that etymologically
distant language pairs are hard to align using mapping approaches.
A non-linear transformation is proposed in \cite{LnmapDeparturMohiud2020}, which
does not assume isomorphism between language pairs, and improved performance on
moderately low-resource languages.
However, \citet{michel-etal-2020-exploring} show that for a very low-resource
language such as Hiligaynon, which has around 300K tokens worth of available
data, good quality monolingual word embeddings cannot be trained, meaning that
they can neither be aligned with other languages.
\citet{eder2021anchorbased} found that mapping approaches on languages under 10M
tokens achieve under 10\% P@1 score when BLI is performed.
In our work, we focus on such low-resource languages and propose to combine
the advantages of related languages in multilingual spaces and hybrid alignment
approaches.

\section{Method}
\label{sec:method}

\begin{figure*}[t]
\begin{center}
\includegraphics[scale=0.6]{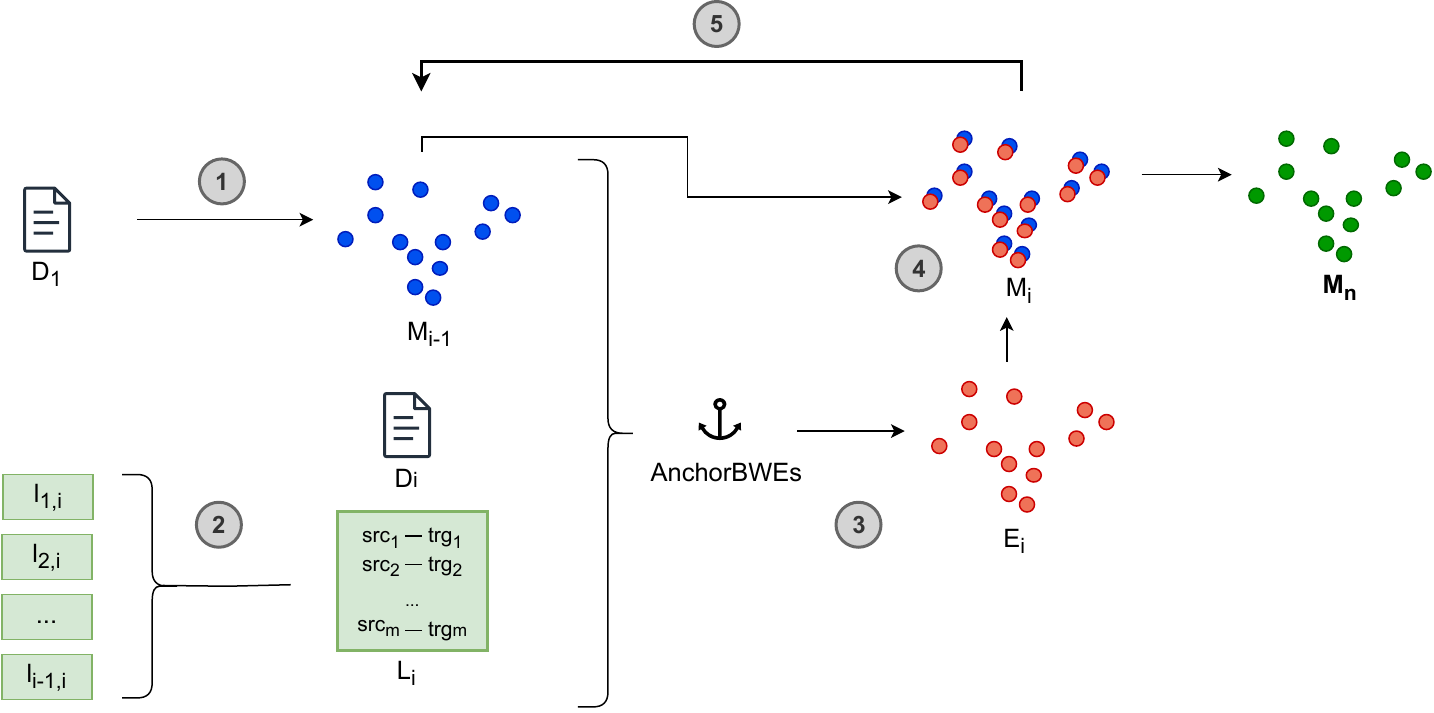}
\caption{Visual depiction of our \ourmethod method.
The resulting embedding ($M_n$ in green) is multilingual involving all languages in the chain.
}
\label{fig:method}
\end{center}
\end{figure*}

The goal of our approach is to reduce the distance between two languages which
are being aligned at a time.
Thus instead of directly aligning the source and target languages we incorporate
a chain of intermediate related languages in order for a reduced distance.
Our approach starts from the source language as the initial multilingual space
and iteratively adds the languages in the chain till it reaches the target language.
We build upon the bilingual \anchors algorithm presented in \citep{eder2021anchorbased}
by extending it to multilingual setting.
First, we discuss the \anchors approach, followed by our proposed
intermediate language-based \ourmethod method.

\subsection{\anchors}

The anchor-based method assumes that the source language is high-resource, thus
starts by training source monolingual word embeddings with a traditional static
word embedding approach, more precisely
\emph{word2vec}~\cite{mikolov2013efficient}.
Using this vector space it trains an embedding space for the low-resource target
language by aligning them
at the same time, this way the
properties of the good quality source space, such as similar embeddings for
words with similar meaning, is transferred to the target space.
Given a seed dictionary defining word translation pairs, the source side of the
pairs are defined as the anchor points.
Instead of randomly initializing all target language words at the beginning of
the training process, the method initializes target words in the seed dictionary
using their related anchor points.
The rest of the training process follows the unchanged algorithm of either CBOW
or Skip-gram on the target language corpus.
This approach significantly outperforms previous methods in low-resource
bilingual
settings, as demonstrated by strong results on both simulated low-resource
language pairs (English-German) and true low-resource language pairs
(English-Hiligaynon).
Additionally, \citet{eder2021anchorbased} shows that not only the cross-lingual
performance is improved, but the monolingual space is of better quality compared
when the target space is trained independently of the source language.

\subsection{\ourmethod}

We extend \anchors by first defining a chain of languages
$C = [c_1, c_2, ..., c_n]$,
starting from the high-resource source language ($c_1$) and ending at the low-resource
target language ($c_n$), including intermediate languages that are related to the
preceding and following nodes.
As described in
Section~\ref{sec:exp}, we define chains in which the
lower-resource languages are of the same language family.
The intuition is to interleave the source and target with languages that are
similar in terms of
linguistic properties.
After selecting the intermediate languages, our method comprises five steps
as depicted in Figure~\ref{fig:method}:
\begin{enumerate}
    \item As the first step ($i=1$), we construct the initial monolingual
    embedding space ($E_1$) for the source language ($c_1$) using its
    monolingual corpus ($D_1$),
        by training a Word2Vec \citep{mikolov2013efficient} model.
    We consider this space as the initial multilingual space ($M_1 := E_1$) which we
    extend in the following steps.
    
    \item In the next step ($i=i+1$), we collect the seed lexicon ($L_i$) for
    training embeddings for the next
    language in the chain ($c_i$) by concatenating the seed lexicons of all the
    languages before $c_i$ in the chain paired with $c_i$.
    More precisely:

    \begin{equation*}
        L_i =  \bigcup_{k=1}^{i-1}l_{k,i}
    \end{equation*}

    \noindent
    where $l_{k,i}$ is the seed lexicon between languages $k$ and $i$.
            Since \citet{eder2021anchorbased} showed that \anchors performs better as
    the number of available anchor points increase, our goal is to take all
    available anchor points already in $M_{i-1}$.

    \item Apply
        \anchors using $M_{i-1}$ as the source embedding space, $D_i$ as the
    training corpus and $L_i$ as the anchors to build
        embeddings ($E_i$) for $c_i$.
                                \item Since \anchors builds embeddings for $c_i$ which are aligned with the
    maintained multilingual space, we simply concatenate them $M_i = M_{i-1} \cup E_i$.
        
    \item Goto step 2 until the target language is reached.

                \end{enumerate}

By strategically integrating intermediate languages, we enrich the quality of
the multilingual space by making sure that the distance between two languages at
any alignment step is minimal.
Our experiments show that without the intermediate languages the quality of the
embeddings built by \anchors is negatively affected by the large gap between the
source and target.

\section{Experimental Setup}
\label{sec:exp}

In this section, we describe the experimental setup, including the selection of
languages, datasets, and model parameters used in our study.

\subsection{Data}

\begin{figure}[t]
\begin{center}
\includegraphics[scale=0.6,trim={2cm 1cm 1cm 1cm},clip]{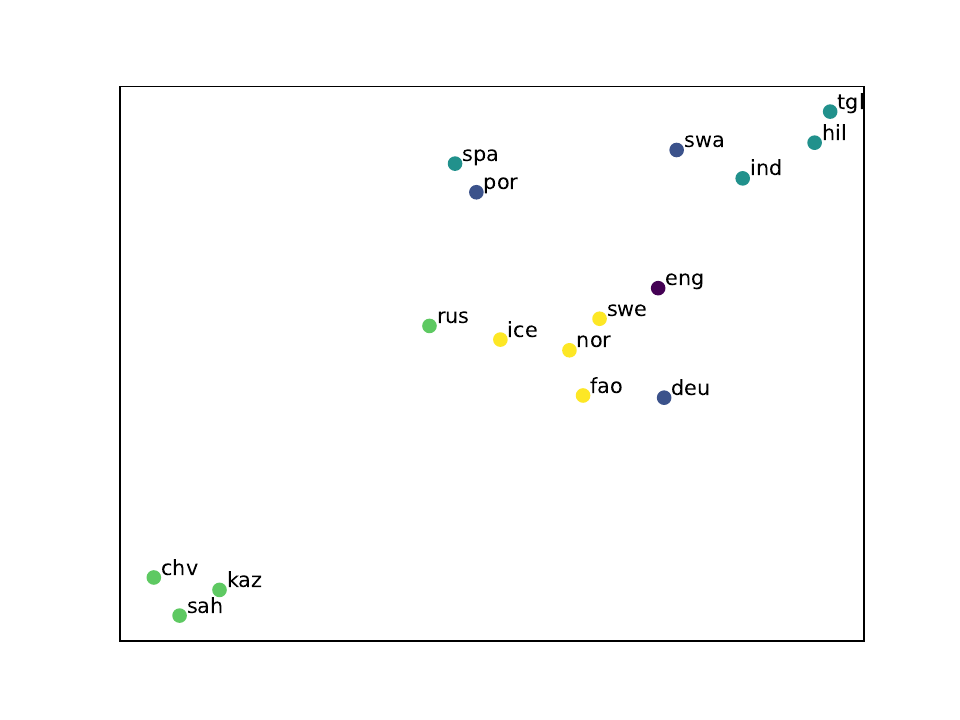}
\caption{
Visualization of language embeddings using \emph{lang2vec} syntax features.
Colors indicate different language families: Austronesian in turquoise, Turkic
in green, Scandinavian in yellow and Atlantic-Congo in blue.
}
\label{fig:langplot}
\end{center}
\end{figure}

We select four language families of different geographic locations for evaluation.
Figure~\ref{fig:langplot} depicts the language similarities in 2D using
\emph{lang2vec} language embeddings based on their syntactic features
\cite{malaviya-etal-2017-learning}.
We discuss their relevance on the final results in Section~\ref{sec:results}.
Although, we selected low-resource target and intermediate languages
based on language families, we stepped over their boundaries in order to have
intermediate languages related to the source language as well by considering
the influence some languages had on others, e.g., during the colonial era.
Our source language is English in each setup,
and sort the intermediate languages based on their monolingual corpora sizes.
We present the exact chains of these languages in section~\ref{sec:results}.

\paragraph{Austronesian}
We select two languages spoken in the Philippines: Tagalog as moderately and
Hiligaynon as very low-resource target languages, with Indonesian and Spanish as
the intermediates.
Spanish being an Indo-European language is related to English.
Additionally, due to colonization, it influenced the selected Austronesian
languages to a varying degree.
Furthermore, Indonesian, Tagalog and Hiligaynon show similarities, especially
the two languages of the Philippines, due to their close proximity.

\paragraph{Turkic}
languages using the Cyrillic script.
We take Kazakh as moderately, and Chuvash and Yakut as very low-resource
languages.
Since they use the Cyrillic alphabet and mostly spoken in Russia, we use Russian
as the intermediate language.
Due to Russian being high-resource, it can be well aligned with English.

\paragraph{Scandinavian}
We select Icelandic and Faroese as two very low-resource languages, with
Norwegian and Swedish as the intermediates that are related to both of
them and to English.

\paragraph{Atlantic-Congo}
Finally, we select Swahili as a moderately low-resource language, which has a
high number of loanwords from Portuguese and German which we take as the
intermediate languages.
We note that we experimented with the very low-resource Zulu and Xhosa languages as well,
however due to difficulties acquiring good quality lexicons for training and
evaluation, we achieved near zero performance, thus we do not present them in this
paper.

\begin{table}[t]
\centering
\small
\begin{tabular}{lllr}
\toprule
& Language & ISO & \# tokens (M)\\
\midrule
\multirow{7}{*}{\rotatebox{90}{intermediate}} & English & eng   & 3~044\\
 & German & deu   & 1~124 \\
 & Spanish & spa & 836 \\
 & Russian & rus   & 717 \\
 & Portuguese & por  & 377 \\
 & Swedish & swe   & 252 \\
 & Indonesian & ind   & 128 \\
 & Norwegian & nor   & 127 \\
\midrule
\multirow{4}{*}{\rotatebox{90}{moderate}} & Kazakh & kaz   & 32 \\
 & Tagalog &  tgl & 11\\
 & Icelandic &  ice & 10 \\
 & Swahili & swa   & 9 \\
\midrule
\multirow{4}{*}{\rotatebox{90}{very-low}} & Chuvash & chv   & 4 \\
 & Yakut & sah   & 3 \\
 & Faroese &  fao & 2 \\
 & Hiligaynon &  hil & 0.35 \\
\bottomrule
\end{tabular}
\caption{
Selected intermediate as well as moderately and very low-resource languages.
Monolingual corpora sizes are shown in millions.
}
\label{table:data}
\end{table}

The embeddings were trained on Wikipedia dumps for all languages except
Hiligaynon, which was trained on the corpus used in
\citep{michel-etal-2020-exploring} due to comparison reasons.
Hiligaynon is extremely low-resource, having 345K tokens in its monolingual
corpus.
Corpus sizes for each language are presented in Table~\ref{table:data}.
Bilingual dictionaries for training and testing are taken from the Wiktionary based
resource released in \cite{AligningWordVIzbick2022}.
As mentioned in the previous section, at each iteration of our approach we take
training dictionaries between the current language and all languages which are
already in the multilingual vector space.
Since, \citet{AligningWordVIzbick2022} only release resources for English paired
with various target languages, we build dictionaries for the other language
pairs through pivoting, more precisely:\begin{multline*}
    l_{k,i} = \{(trg_{e,k}, trg_{e,i}) \mid \\
    (src_{e,k}, trg_{e,k}, src_{e,i}, trg_{e,i}) \in l_{e,k}\times l_{e,i}, \\
    src_{e,i}=src_{e,k}\}
\end{multline*}

\noindent
where $l_{e,x}$ is a dictionary between English ($e$) and an arbitrary language
($x$), while $src_{x,y}$ and $trg_{x,y}$ is a source ($x$) and target ($y$)
language translation pair.
Number of dictionary entries for each language pair is presented in
Table~\ref{table:dicosize}.

\begin{table}[t]
\centering
\resizebox{.84\columnwidth}{!}{\begin{tabular}{lrr|lr}
lang.  & train            & test            & lang.  & train             \\ \hline
eng-deu  & \numprint{65120} & -               & spa-ind  & \numprint{19952} \\
eng-spa  & \numprint{88114} & -               & spa-tgl  & \numprint{26088} \\
eng-rus  & \numprint{67397} & -               & spa-hil & \numprint{4661}  \\
eng-por  & \numprint{53336} & -               & rus-kaz  & \numprint{21147} \\
eng-swe  & \numprint{25214} & -               & rus-chv  & \numprint{1212}  \\
eng-ind  & \numprint{9868}  & -               & rus-sah & \numprint{6913}  \\
eng-nor  & \numprint{18916} & -               & por-swa & \numprint{13197} \\
eng-kaz  & \numprint{8990}  & \numprint{2358} & swe-nor  & \numprint{15843} \\
eng-tgl  & \numprint{15242} & \numprint{2597} & swe-ice  & \numprint{13749} \\
eng-ice  & \numprint{17004} & \numprint{2568} & swe-fao  & \numprint{6425}  \\
eng-swa & \numprint{5203}  & \numprint{2132} & ind-tgl  & \numprint{6089}  \\
eng-chv  & \numprint{170}   & \numprint{823}  & ind-hil & \numprint{1575}  \\
eng-sah & \numprint{1202}  & \numprint{2065} & nor-ice  & \numprint{10759} \\
eng-fao  & \numprint{4505}  & \numprint{1786} & nor-fao  & \numprint{4917}  \\
eng-hil & \numprint{1132}  & \numprint{200}  & kaz-chv  & \numprint{160}   \\
deu-por  & \numprint{44791} & -               & kaz-sah & \numprint{1000}  \\
deu-swe  & \numprint{34659} & -               & tgl-hil & \numprint{1683}  \\
deu-swa & \numprint{14818} & -               & ice-fao  & \numprint{5587}
\end{tabular}}
\caption{Number of unique words in the train and test dictionaries of the used
language pairs.}
\label{table:dicosize}
\end{table}

\subsection{Baselines and Model Parameters}

We compare our approach to the mapping-based bilingual \emph{VecMap}
\cite{artetxe-etal-2018-robust} and multilingual \emph{UMWE}
\cite{chen-cardie-2018-unsupervised} approaches.
Additionally, we run \anchors \cite{eder2021anchorbased} as our joint alignment
baseline.

We trained word2vec  embeddings \citep{mikolov2013efficient} with a maximum
vocabulary size of \numprint{200000} in every setup, i.e., for the mapping-based
baselines as well as in \anchors and \ourmethod.
The training was performed using standard hyperparameters included in the
{Gensim} Word2Vec package \citep{rehurek_lrec}: context window of 5,
dimensionality of 300 and for 5 epochs, with the exception that we used minimum
word frequency of 3 due to the small corpora for the target languages.
Additionally, since \citet{eder2021anchorbased} showed that CBOW
outperforms SG in \anchors, we used the former in our experiments.

We use the MUSE evaluation tool
\citep{conneau2018word}
to report precision at 1, 5, and 10, using the nearest neighbor search.
For the mapping based approaches we leverage the CSLS similarity score
as it was shown to perform better by handling the hubness
problem \cite{conneau2018word}.
However, similarly to \cite{woller-etal-2021-neglect} we found that jointly
trained embeddings do not benefit from the CSLS method, thus we use simple
cosine similarity (NN) based search for both \anchors and \ourmethod.

\section{Results}
\label{sec:results}

\begin{table*}[h!]
\centering
\resizebox{.60\linewidth}{!}{
\begin{tabular}{lllrrrr}
\toprule
& Method & Intermediate & P@1  &  P@5 & P@10
 \\
\toprule
\multicolumn{6}{c}{Moderately low-resource} \\
\midrule
\multirow{4}{*}{\rotatebox{90}{Kazakh}} & VecMap     & -   & 12.37          & 23.06          & 29.42 \\
                                        & UMWE       & rus & \textbf{14.58} & 25.18          & 29.95  \\
                                        & \anchors   & -   & 12.79          & 24.51          & 31.22  \\
                                        & \ourmethod & rus & {14.37}        & \textbf{26.90} & \textbf{33.16} \\
\midrule

\multirow{4}{*}{\rotatebox{90}{Tagalog}} & VecMap     & -         & 7.63           & 14.94          & 17.76 \\
                                         & UMWE       & esp - ind & 15.59          & 24.69          & 29.08 \\
                                         & \anchors   & -         & 15.38          & 26.57          & 32.01 \\
                                         & \ourmethod & esp - ind & \textbf{15.90} & \textbf{28.66} & \textbf{33.79} \\
\midrule

\multirow{4}{*}{\rotatebox{90}{Icelandic}} & VecMap     & -         & 4.48           & 9.26           & 12.68 \\
                                           & UMWE       & swe - nor & \textbf{12.35} & 18.23          & 21.02 \\
                                           & \anchors   & -         & 8.77           & 17.94          & 21.67 \\
                                           & \ourmethod & swe - nor & 8.17           & \textbf{18.75} & \textbf{23.19}\\
\midrule

\multirow{4}{*}{\rotatebox{90}{Swahili}} & VecMap     & -         & 2.29           & 7.08           & 10.68 \\
                                         & UMWE       & deu - por & \textbf{13.38} & \textbf{24.05} & \textbf{28.07}  \\
                                         & \anchors   & -         & 10.23          & {21.44}        & { 26.22} \\
                                         & \ourmethod & deu - por & {10.99}        & 20.78          & 25.90 \\
\bottomrule
\multicolumn{6}{c}{Very low-resource} \\
\midrule

\multirow{4}{*}{\rotatebox{90}{Chuvash}} & VecMap     & -   & 0.00           & 0.00          & 0.00 \\
                                         & UMWE       & rus & 0.00           & 0.30          & 0.30 \\
                                         & \anchors   & -   & \textbf{ 0.31} & 0.61          & 1.53 \\
                                         & \ourmethod & rus & \textbf{0.31}  & \textbf{0.92} & \textbf{2.75} \\
\midrule

\multirow{4}{*}{\rotatebox{90}{Yakut}}  & VecMap     & -   & 0.00          & 0.25          & 0.38 \\
                                        & UMWE       & rus & 0.76          & 1.78          & 2.42 \\
                                        & \anchors   & -   & \textbf{2.92} & \textbf{7.49} & \textbf{9.90}  \\
                                        & \ourmethod & rus & 2.03          & 6.98          & 9.14\\
\midrule

\multirow{4}{*}{\rotatebox{90}{Faroese}} & VecMap     & -         & 0.00          & 0.51          & 0.63 \\
                                         & UMWE       & swe - nor & 1.01          & 3.42          & 3.93  \\
                                         & \anchors   & -         & 4.09          & 9.20          & 12.26\\
                                         & \ourmethod & swe - nor & \textbf{4.21} & \textbf{9.96} & \textbf{	13.67} \\

\midrule

\multirow{4}{*}{\rotatebox{90}{Hiligaynon}} & VecMap     & -               & 0.00          & 0.00          & 0.00 \\
                                            & UMWE       & esp - ind       & 0.00          & 0.00          & 0.00 \\
                                                                                                                                    & \anchors   & -               & \textbf{5.08}          & \textbf{7.63} & {8.47}\\
                                            & \ourmethod & esp - ind       & \textbf{5.08}          & 6.78          & \textbf{10.17}\\
                                                                                        
\bottomrule
\end{tabular}
}
\caption{
Precision at $k \in \{1, 5, 10\}$ values for the target languages
        paired with English as the source in each case.
        The \emph{Intermediate} column shows the languages in between the source and target (e.g., line 2 shows the chain
        \textit{\underline{English}$\rightarrow$Russian$\rightarrow$\underline{Kazakh}}
}
\label{table:results}
\end{table*}

We present our results in Table~\ref{table:results} split into the moderately
and very low-resource language groups and sorted based on the size of available
monolingual data for each target language (Table~\ref{table:data}).
Overall, the results show the difficulties of building cross-lingual word
embeddings for the selected target languages, since the performance is much
lower compared to high resource languages in general, which for example is around
50\% P@1 for English-German on the Wiktionary evaluation set
\cite{AligningWordVIzbick2022}.
Comparing the multilingual UMWE approach to the bilingual VecMap the results
support the use of related languages, since they improve the performance on most
source-target language pairs.
However, this is most apparent on the moderately low-resource languages.
The results on the very low-resource languages are very poor for the
mapping-based approaches, which as discussed depend on the quality of
pre-trained monolingual embeddings.
In contrast, the semi-joint anchor-based approaches can significantly improve the
embedding quality showing their superiority in the very low-resource setups.

Our proposed \ourmethod method outperforms mapping-based approaches on 7 out of
8 target languages, and \anchors on 6 target
languages, which is most apparent when retrieving more than one translation
candidate (P@5 and P@10).
Interestingly when looking at P@1, the systems are close to each other,
indicating that our method improves the general neighborhood relations of the
embedding space instead of just improving the embeddings of a few individual
words.
This is further supported in the case of Kazakh and Icelandic where UMWE outperforms
\ourmethod in terms of P@1, however it performs lower when a larger neighborhood
is leveraged for the translation.
This property is caused by the combination of the semi-joint anchor-based
training, instead of relying on independently trained monolingual spaces, and
the smaller distances between aligned languages.

When comparing moderately and very low-resource languages, we found similar
trends in the two groups.
In both cases \ourmethod outperforms \anchors on 3 out of 4 languages, however
in case of Hiligaynon, which has less than 1 million tokens, the results are
mixed, i.e., \anchors tends to perform better when the smaller neighborhood of
P@5 is considered, but it is the opposite when P@10 is measured.
Furthermore, UMWE tends to be more competitive with \anchors on the moderately
low-resource languages, e.g., it performs
better in case of Kazakh, while it does not improve over \ourmethod.
Overall however, we found no strong correlation between the available monolingual
resources for a given language and on which target language \ourmethod achieved
the best results, since the two cases where it did not improve over the
baselines are the $3^{rd}$ (Yakut) and $5^{th}$ (Swahili) lowest resource languages.
Looking at the visualization of language embeddings in
Figure~\ref{fig:langplot}, the negative results on Swahili can be explained by
the relatively large distance between its two intermediate pairs.
Although Swahili has a large number of German and Portuguese loan words, the
syntactic properties of the languages seem to be too different.
Similarly, Yakut (sah) is the furthest away from Russian which could explain our
negative results.

\begin{table}[t]
\centering
\resizebox{.99\columnwidth}{!}{
\begin{tabular}{lllrrrr}
\toprule
& Method & Inter. & P@1  &  P@5 & P@10
 \\
\midrule

\multirow{2}{*}{\rotatebox{90}{sah}}    & \ourmethod & rus & 2.03          & 6.98          & 9.14\\
                                        & \ourmethod & rus - kaz & 1.78          & 5.58          & 8.12\\
\midrule

\multirow{2}{*}{\rotatebox{90}{fao}}      & \ourmethod & swe - nor & {4.21} & {9.96} & {	13.67} \\
                                          & \ourmethod & swe - nor - ice & 3.83 & 7.15 & 8.81 \\

\midrule

\multirow{2}{*}{\rotatebox{90}{hil}}         & \ourmethod & esp - ind & {5.08} & 6.78 & {10.17}\\
                                             & \ourmethod & esp - ind - tgl & {5.08} & 6.78 & {7.63}\\

\bottomrule
\end{tabular}
}
\caption{
Experiments on adding related moderately low-resource languages to the language
chains of very low-resource languages.
}
\label{table:moderate}
\end{table}

\subsection{Adding Moderate Resource Languages}

Since some moderately low-resource languages are related to the very
low-resource ones (Kazakh to Yakut\footnote{Kazakh is also related to Chuvash
which we omitted in these experiments due to low results on Chuvash in
general.}, Icelandic to Faroese and Tagalog to Hiligaynon), we add them to the
language chain in the experiments presented in Table~\ref{table:moderate}.
The results show, that although these languages are closely related, they do not
contribute positively to the quality of the resulting MWEs.
These results indicate, that the languages involved in the language-chains as
intermediate steps should have good quality embeddings (the BLI performance P@5
for the Russian, Swedish, Norwegian and Spanish range between 45\% and 65\%), thus
embedding quality is more important than language closeness.
Additionally, Figure~\ref{fig:langplot} shows that Tagalog is less similar to
Indonesian and Spanish than to Hiligaynon, and Icelandic is less similar to Faroese
than to Norwegian or Swedish.

\begin{table}[t]
\centering
\resizebox{.99\columnwidth}{!}{
\begin{tabular}{lllrrrr}
\toprule
& Method & Inter. & P@1  &  P@5 & P@10
 \\
\toprule
\multicolumn{6}{c}{Moderately low-resource} \\
\midrule
\multirow{2}{*}{\rotatebox{90}{kaz}} & \ourmethod & rus & {14.37}        & {26.90} & {33.16} \\
                                        & \textsc{ChainMWEs}$^*$ & rus & {13.67} & {26.19} & {31.22} \\
\midrule

\multirow{2}{*}{\rotatebox{90}{tgl}} & \ourmethod & esp - ind & {15.90} & {28.66} & {33.79} \\
                                         & \textsc{ChainMWEs}$^*$ & esp - ind & {13.28} & {23.43} & {28.66} \\
\midrule

\multirow{2}{*}{\rotatebox{90}{ice}} & \ourmethod & swe - nor & 8.17           & {18.75} & {23.19}\\
                                           & \textsc{ChainMWEs}$^*$ & swe - nor & 8.27           & {15.42} & {19.96}\\
\midrule

\multirow{2}{*}{\rotatebox{90}{swa}} & \ourmethod & deu - por & {10.99}        & 20.78          & 25.90 \\
                                         & \textsc{ChainMWEs}$^*$ & deu - por & {11.21}        & 20.67          & 24.92 \\
\bottomrule
\multicolumn{6}{c}{Very low-resource} \\
\midrule

\multirow{2}{*}{\rotatebox{90}{chv}} & \ourmethod & rus & {0.31}  & {0.92} & {2.75} \\
                                         & \textsc{ChainMWEs}$^*$ & rus & {0.61}  & {1.53} & {3.67} \\
\midrule

\multirow{2}{*}{\rotatebox{90}{sah}}  & \ourmethod & rus & 2.03          & 6.98          & 9.14\\
                                        & \textsc{ChainMWEs}$^*$ & rus & 2.28          & 6.85          & 9.01 \\
\midrule

\multirow{2}{*}{\rotatebox{90}{fao}} & \ourmethod & swe - nor & {4.21} & {9.96} & {	13.67} \\
                                         & \textsc{ChainMWEs}$^*$ & swe - nor & {3.96} & {8.56} & {12.52} \\

\midrule

\multirow{2}{*}{\rotatebox{90}{hil}} & \ourmethod & esp - ind       & {5.08}          & 6.78          & {10.17}\\
                                            & \textsc{ChainMWEs}$^*$ & esp - ind       & {4.24}          & 5.93           & {8.47} \\

\bottomrule
\end{tabular}
}
\caption{
Results of the ablation experiments, where we turn training dictionary
accumulation off in \textsc{ChainMWEs}$^*$, by using only the dictionary between a given language and its preceding neighbor.
}
\label{table:ablation}
\end{table}

\subsection{Ablation Study}

An advantage of the sequential nature of our approach is that as we add more
languages to the multilingual space step-by-step, the number of potential
anchor points for aligning the language next in line increases.
We exploit this by accumulating all word translation pairs from the dictionaries
between all languages already in the multilingual space and the currently
trained language (Step 2).
Although this requires dictionaries between all language pairs, we mitigated
this requirement by pivoting through English.
In Table~\ref{table:ablation} we present an ablation study, where we turn
dictionary accumulation off, by using dictionaries only between the trained
language and its preceding neighbor.
The results show that this has a sizable impact on the performance.
Although there are a few cases where P@1 is marginally improved (Icelandic,
Swahili, Chuvash and Yakut), both P@5 and P@10 are decreased in most cases even
where P@1 is improved except Chuvash.
The least impacted by the accumulated dictionaries are Turkic languages which
indicates their strong relation to Russian and distance from English which could
stem from their different scripts.
Overall, these findings align with the results of \cite{eder2021anchorbased}, who
showed that the embedding quality improves as more dictionary entries are
available.

\section{Conclusion}
\label{sec:conclusions}

In this paper we proposed \ourmethod, a novel method for enhancing multilingual embeddings
of low-resource languages by incorporating intermediate languages to bridge the
gap between distant source and target languages.
Our approach extends \anchors, the bilingual approach of \citet{eder2021anchorbased} to
MWEs by employing chains of related languages.
We evaluate \ourmethod on 4 language families involving 4 moderately and 4 very
low-resource languages using bilingual lexicon induction.
Our results demonstrate the effectiveness of our method showing improvements on
6 out of 8 target languages compared to both bilingual and multilingual
mapping-based, and the \anchors baselines.
Additionally, we show the importance of involving only those intermediate languages for which
building good quality embeddings is possible.

\section*{Limitations}

One limitation of our work is the manual selection of intermediate languages.
Although,
the selection and ordering of languages in the chains was straightforward
based on language family information, such as Glottolog
\cite{nordhoff2011glottolog},
and available data size,
it could be possible that
other languages which we did not consider in our experiments are also helpful in
improving the quality of MWEs.
Additionally, we did not consider all possible ordering of intermediate
languages, such as the order of
English$\rightarrow$Norwegian$\rightarrow$Swedish$\rightarrow$Faroese instead of
English$\rightarrow$Swedish$\rightarrow$Norwegian$\rightarrow$Faroese, in order
to save resources.
Thus, a wider range of chains could uncover further improvements.

\section*{Acknowledgements}

We thank
the anonymous reviewers for their helpful feedback and
the Cambridge
LMU Strategic Partnership for
funding for this
project.\footnote{\url{https://www.cambridge.uni-muenchen.de}}
The work was also funded by the European Research Council
(ERC; grant agreements No.~740516 and No.~640550)
and by the German Research Foundation (DFG; grant FR 2829/4-1).

\bibliography{anthology,custom}
\bibliographystyle{acl_natbib}

\end{document}